\documentclass{article}

 \usepackage[preprint]{neurips_2025}

\usepackage[utf8]{inputenc} 
\usepackage[T1]{fontenc}    
\usepackage{hyperref}       
\usepackage{url}            
\usepackage{booktabs}       
\usepackage{amsfonts}       
\usepackage{amsthm}
\usepackage{nicefrac}       
\usepackage{microtype}      
\usepackage{xcolor}         
\usepackage{amsmath}
\usepackage{cleveref}
\usepackage{mathtools}
\usepackage{caption}
\usepackage{subcaption}
\usepackage{adjustbox}
\usepackage{cancel}
\usepackage{enumitem}

\usepackage{algorithm}
\usepackage{float}
\usepackage[noEnd=true, indLines=true]{algpseudocodex}
\algrenewcommand\algorithmicrequire{\textbf{Inputs:}}
\algrenewcommand\algorithmicensure{\textbf{Outputs:}}

\usepackage{amsmath,amsfonts,bm}

\def\eqref#1{equation~\ref{#1}}

\def\1{\bm{1}}

\def\vzero{{\bm{0}}}

\def\vx{{\bm{x}}}

\def\mI{{\bm{I}}}

\DeclareMathAlphabet{\mathsfit}{\encodingdefault}{\sfdefault}{m}{sl}
\SetMathAlphabet{\mathsfit}{bold}{\encodingdefault}{\sfdefault}{bx}{n}

\def\gA{{\mathcal{A}}}

\def\gE{{\mathcal{E}}}

\def\gL{{\mathcal{L}}}
\def\gM{{\mathcal{M}}}
\def\gN{{\mathcal{N}}}

\def\gS{{\mathcal{S}}}

\newcommand{\E}{\mathbb{E}}

\newcommand{\KL}{D_{\mathrm{KL}}}

\DeclareMathOperator*{\argmax}{arg\,max}

\input{style/widebar}

\crefformat{chapter}{#2Chapter~#1#3}
\crefmultiformat{chapter}{#2Chapters~#1#3}%
{ \&~#2#1#3}{, #2#1#3}{, \&~#2#1#3}
\crefformat{section}{#2Section~#1#3}
\crefmultiformat{section}{#2Sections~#1#3}%
{ \&~#2#1#3}{, #2#1#3}{, \&~#2#1#3}
\crefformat{appendix}{#2Appendix~#1#3}

\crefformat{equation}{(#2#1#3)}
\crefmultiformat{equation}{#2Equations~#1#3}%
{ \&~#2#1#3}{, #2#1#3}{, \&~#2#1#3}

\crefformat{table}{#2Table~#1#3}
\crefformat{figure}{#2Figure~#1#3}
\crefmultiformat{figure}{#2Figures~#1#3}%
{ \&~#2#1#3}{, #2#1#3}{, \&~#2#1#3}

\crefformat{theorem}{#2Theorem~#1#3}
\crefmultiformat{theorem}{#2Theorems~#1#3}%
{ \&~#2#1#3}{, #2#1#3}{, and~(#2#1#3)}

\crefformat{lemma}{#2Lemma~#1#3}
\crefformat{proposition}{#2Proposition~#1#3}
\crefmultiformat{proposition}{#2Propositions~#1#3}%
{ \&~#2#1#3}{, #2#1#3}{, and~(#2#1#3)}

\crefformat{algorithm}{#2Algorithm~#1#3}
\crefmultiformat{algorithm}{#2Algorithms~#1#3}%
{ \&~#2#1#3}{, #2#1#3}{, and~(#2#1#3)}

\crefformat{corollary}{#2Corollary~#1#3}
\crefformat{definition}{#2Definition~#1#3}
\crefformat{remark}{#2Remark~#1#3}
\crefformat{example}{#2Example~#1#3}

\crefformat{assumption}{#2Assumption~#1#3}
\crefmultiformat{assumption}{#2Assumptions~#1#3}%
{ \&~#2#1#3}{, #2#1#3}{, and~(#2#1#3)}

\title{Relative Trajectory Balance is equivalent to Trust-PCL}

\newcommand{\ie}{\textit{i.e.},~}
\newcommand{\eg}{\textit{e.g.},~}
\newcommand{\prior}{\mathrm{prior}}
\newcommand{\terminal}{\bot}
\renewcommand{\KL}{\mathrm{KL}}

\newcommand{\soft}{\mathrm{soft}}

\newcommand{\trustpcl}{\mathrm{T\textrm{-}PCL}}

\newtheorem{proposition}{Proposition}

\definecolor{mydarkblue}{rgb}{0,0.08,0.45}
\hypersetup{ %
    colorlinks=true,
    linkcolor=mydarkblue,
    citecolor=mydarkblue,
    filecolor=mydarkblue,
    urlcolor=mydarkblue,
}

\makeatletter
\def\thanks#1{\protected@xdef\@thanks{\@thanks
        \protect\footnotetext{#1}}}
\makeatother

\author{%
    Tristan~Deleu\:\!{\normalfont\textsuperscript{1}}\thanks{\textsuperscript{1}Entalpic, \textsuperscript{2}McGill University,  \textsuperscript{3}Universit\'{e} de Montr\'{e}al, \textsuperscript{4}Google DeepMind.}\thanks{\textsuperscript{\phantom{1}}Correspondence: Tristan Deleu (\href{mailto:deleutri@mila.quebec}{deleutri@mila.quebec})\\[0.2ex]\hspace*{2.155em}Code: \href{https://github.com/padideee/rtb-trustpcl-finetuning}{https://github.com/padideee/rtb-trustpcl-finetuning}} \qquad Padideh~Nouri\:\!{\normalfont\textsuperscript{2}} \qquad Yoshua~Bengio\:\!{\normalfont\textsuperscript{3}} \qquad Doina~Precup\:\!{\normalfont\textsuperscript{2,4}}\\[1ex]
    Mila -- Quebec AI Institute
}

\begin{document}

\maketitle

\begin{abstract}
    Recent progress in generative modeling has highlighted the importance of Reinforcement Learning (RL)  for fine-tuning, with KL-regularized methods in particular proving to be highly effective for both autoregressive and diffusion models. Complementing this line of work, the Relative Trajectory Balance (RTB) objective was recently introduced in the context of Generative Flow Networks (GFlowNets) to serve the same role of improving fine‑tuning in sequential generative models. Building on prior work linking GFlowNets and maximum‑entropy RL, we establish in this paper an equivalence between RTB and Trust-PCL, an off-policy RL method with KL regularization. This equivalence situates RTB within the broader theoretical landscape of KL-regularized RL, and clarifies its relationship to earlier methods. Leveraging this insight, we revisit an illustrative example from the RTB paper and show that KL-regularized RL methods achieve comparable performance, offering an alternative perspective to what was previously reported.
\end{abstract}

\section{Introduction}
\label{sec:introduction}
Generative models are playing an increasingly prominent role in artificial intelligence, ranging from language and reasoning \citep{brown2020gpt3,wei2022cot}, to vision \citep{ho2020ddpm,rombach2022stablediffusion} and scientific discovery \citep{abramson2024alphafold3}. At the heart of this rapid progress, fine-tuning is central to aligning models with downstream applications while preserving diversity and prior behavior. To complement supervised fine-tuning \citep{ouyang2022rlhf}, Reinforcement Learning (RL) with KL regularization has proven effective and is widely used for this task \citep{jaques2017sequencetutor,uehara2024rlfinetuningdiffusion}, where reward maximization is traded off against maintaining consistency with a pretrained model. In parallel, Generative Flow Networks \citep[GFlowNets;][]{bengio2021gflownet,bengio2023gflownetfoundations} have been proposed as an alternative paradigm to train generative models with techniques inspired by RL. They have since gained popularity in areas such as combinatorial optimization \citep{zhang2023graphcogfn}, causal discovery \citep{deleu2022daggflownet}, and scientific discovery \citep{jain2023gfnscientific}.

Extending ideas from GFlowNets, the \emph{Relative Trajectory Balance} objective \citep[RTB;][]{venkatraman2024rtb} was recently proposed as an alternative to KL-regularized RL for fine-tuning generative models, framing the problem in terms of Trajectory Balance \citep{malkin2022trajectorybalance}. In this paper, we show that RTB is exactly equivalent to an existing RL algorithm called \emph{Trust-PCL} \citep{nachum2018trustpcl}, reinforcing the strong ties existing between GFlowNets and (entropy-regularized) RL \citep{tiapkin2024gfnmaxentrl,mohammadpour2024maxentgfn,deleu2024gfnmaxentrl,jiralerspong2025tgm}. In particular, while \citet{venkatraman2024rtb} suggested that KL-regularized RL methods may perform poorly even on simple tasks, we argue and show on their illustrative example that these issues stem from algorithmic and reward choices rather than from any fundamental limitation of RL with KL regularization itself. This new perspective places RTB within the well-established framework of KL-regularized RL and underscores its actual source of effectiveness.

\section{Background}
\label{sec:background}
We study the problem of training a \emph{sequential generative model} capable of sampling objects as a sequence of multiple steps. This covers, for example, the case of diffusion samplers \citep{sendera2024diffusiongfn} that generate images $\vx_{T}$ as a sequence of denoising steps $\vx_{1} \rightarrow \vx_{2} \rightarrow \ldots \rightarrow \vx_{T}$, starting from some Gaussian noise $\vx_{1} \sim \gN(\vzero, \mI)$ and sequentially sampling from a denoising model $\vx_{t+1} \sim P_{\phi}(\vx_{t+1}\mid \vx_{t})$. By following this iterative process, we eventually obtain samples from the \emph{marginal distribution} defined by
\begin{equation}
    P_{\phi}^{\top}(\vx_{T}) = \int P_{\phi}(\vx_{1}, \vx_{2}, \ldots, \vx_{T})\, d\vx_{1:T-1} = \int \bigg[P(\vx_{1})\prod_{t=1}^{T-1}P_{\phi}(\vx_{t+1}\mid \vx_{t})\bigg]\,d\vx_{1:T-1}.
    \label{eq:terminating-state-distribution}
\end{equation}
We assume that we have access to a base sequential generative model, with transition probabilities $\pi_{\prior}(\vx_{t+1}\mid \vx_{t})$, such as a model pre-trained with a large dataset of observations, and that we are also given a fixed \emph{energy function} $\gE(\vx_{T})$ that we can query for any object generated by the model. Our objective in this paper is to learn a sequential generative model $P_{\phi}(\vx_{t+1}\mid \vx_{t})$ such that its marginal distribution is the \emph{tilted distribution}, modulated by $\gE(\vx_{T})$:
\begin{equation}
    P_{\phi}^{\top}(\vx_{T}) \propto \pi_{\prior}^{\top}(\vx_{T})\exp(-\gE(\vx_{T})/\alpha),
    \label{eq:tilted-distribution}
\end{equation}
for some temperature parameter $\alpha > 0$. The role of the energy function is to steer generation towards samples with desired properties (\eg generating images based on a description).

\subsection{Entropy-regularized Reinforcement Learning}
\label{sec:entropy-regularized-rl}
One way to approach this problem is to view it from the perspective of Reinforcement Learning \citep{korbak2022rlwithklbayesian}. We consider a finite-horizon Markov Decision Process (MDP) $\gM = (\widebar{\gS}, \gA, r)$, where $\widebar{\gS} = \gS \cup \{\terminal\}$ is a state space augmented by a special state $\terminal \notin \gS$ indicating the end of a trajectory, $\gA$ is the action space, and $r$ is a reward function that we will detail below. We identify a state $s_{0} \in \gS$ called the \emph{initial state} from which all trajectories start; these trajectories are guaranteed to end in $\terminal$ since the MDP is finite-horizon. To give a concrete example, we may see the multiple steps of denoising in a diffusion sampler as an MDP with a specific structure \citep{fan2023dpok,black2024ddpo}: its states are of the form $s_{t} = (\vx_{t}, t)$, transitions (actions) correspond to applying one step of denoising $(\vx_{t}, t) \rightarrow (\vx_{t+1}, t+1)$, and trajectories must terminate when a state of the form $(\vx_{T}, T)$ is reached after $T$ steps (termination being indicated by the transition $(\vx_{T}, T) \rightarrow \terminal$).

Following \citet{deleu2024gfnmaxentrl}, the reward function $r(s, s')$ obtained when transitioning from $s\rightarrow s'$ is defined such that the sum of rewards along a trajectory only depends on the energy of the final state it reaches right before terminating; in other words, for a trajectory $\tau = (s_{0}, s_{1}, \ldots, s_{T}, \terminal)$, we have
\begin{equation}
    \sum_{t=0}^{T}r(s_{t}, s_{t+1}) = -\gE(s_{T}),
    \label{eq:return-energy}
\end{equation}
where we will use the convention $s_{T+1} = \terminal$ throughout this paper. This includes, in particular, the case where the reward is only obtained at the end of the trajectory (\ie $r(s_{T}, \terminal) = -\gE(s_{T})$, and zero everywhere else, also called \emph{outcome-based reward}; \citealp{uesato2022ormprm}).

Contrary to standard RL, where the objective is to find a policy $\pi(s_{t+1}\mid s_{t})$ that maximizes the expected sum of rewards, \emph{KL-regularized RL} includes an additional KL regularization term between the current policy and an anchor policy $\pi_{\prior}$ we don't want to deviate too much from:
\begin{equation}
    \pi^{\star}_{\mathrm{RelEnt}} = \argmax_{\pi} \E_{\tau \sim \pi}\left[\sum_{t=0}^{T}r(s_{t}, s_{t+1}) - \alpha \KL\big(\pi(\cdot\mid s_{t})\,\|\,\pi_{\prior}(\cdot\mid s_{t})\big)\right].
    \label{eq:kl-regularized-rl-objective}
\end{equation}
It can be shown that with the particular choice of reward function made in \cref{eq:return-energy}, the marginal distribution associated with this optimal policy is the tilted distribution \citep{nachum2018trustpcl,korbak2022rlwithklbayesian}
\begin{equation}
    \pi^{\star\top}_{\mathrm{RelEnt}}(s_{T}) \propto \pi_{\prior}^{\top}(s_{T})\exp(-\gE(s_{T})/\alpha),
    \label{eq:tilted-optimal-kl-rl}
\end{equation}
where the temperature parameter $\alpha$ naturally emerges from the regularization constant in \cref{eq:kl-regularized-rl-objective}; a proof of this is given in \cref{app:marginal-distribution-optimal-policy} for completeness. This constitutes the foundation of the recent line of work using RL for fine-tuning language models \citep{jaques2017sequencetutor,ouyang2022rlhf} and diffusion models \citep{fan2023dpok,black2024ddpo,uehara2024rlfinetuningdiffusion}.

\subsection{Relative trajectory balance}
\label{sec:relative-trajectory-balance}
Taking another perspective, this time rooted in the literature on GFlowNets \citep{bengio2023gflownetfoundations}, \citet{venkatraman2024rtb} introduced an objective called the \emph{Relative Trajectory Balance} (RTB) loss to sample from the tilted distribution \cref{eq:tilted-distribution}. Given a transition probability $P_{\phi}(s_{t+1}\mid s_{t})$ and a scalar $Z_{\psi} > 0$, the RTB loss is a non-linear least-square objective $\gL_{\mathrm{RTB}}(\phi, \psi) = \frac{1}{2}\E_{\pi_{b}}[\Delta_{\mathrm{RTB}}^{2}(\tau; \phi, \psi)]$, where $\pi_{b}$ is an arbitrary distribution over trajectories (this is an \emph{off-policy} objective), and the residual is defined as
\begin{equation}
    \Delta_{\mathrm{RTB}}(\tau; \phi, \psi) = \log \frac{\prod_{t=0}^{T}\pi_{\prior}(s_{t+1}\mid s_{t})}{Z_{\psi}\prod_{t=0}^{T}P_{\phi}(s_{t+1}\mid s_{t})} - \frac{\gE(s_{T})}{\alpha}.
    \label{eq:relative-trajectory-balance-residual}
\end{equation}
\citet{venkatraman2024rtb} showed that if this RTB loss is minimized, then the marginal distribution associated with the optimal $P_{\phi}(s_{t+1}\mid s_{t})$ satisfies \cref{eq:tilted-distribution}, and the optimal $Z_{\psi}$ is the normalization constant of this tilted distribution.

\section{Equivalence between RTB and Trust-PCL}
\label{sec:equivalence-rtb-trust-pcl}
In line with the growing body of work recently connecting GFlowNets with Maximum Entropy Reinforcement Learning (MaxEnt-RL) \citep{tiapkin2024gfnmaxentrl,mohammadpour2024maxentgfn,deleu2024gfnmaxentrl}, where all the losses introduced in the GFlowNet literature have been shown to have a counterpart in MaxEnt-RL, one may wonder whether the RTB loss can also be viewed from the point of view of RL. In fact, we saw that solving \cref{eq:kl-regularized-rl-objective} already provides a way to sample from the tilted distribution \cref{eq:tilted-optimal-kl-rl}, and there exists an off-policy algorithm called \emph{Trust-PCL} \citep{nachum2018trustpcl} that does exactly this. For a policy $\pi_{\phi}(s_{t+1}\mid s_{t})$ and a soft state-value function $V_{\soft}^{\psi}(s)$, Trust-PCL is also defined as a non-linear least-square objective $\gL_{\trustpcl}(\phi, \psi) = \tfrac{1}{2}\E_{\pi_{b}}[\Delta_{\trustpcl}^{2}(\tau;\phi, \psi)]$, where
\begin{equation}
\Delta_{\trustpcl}(\tau; \phi, \psi) = -V_{\soft}^{\psi}(s_{0}) + \sum_{t=0}^{T}r(s_{t}, s_{t+1}) + \alpha \log \frac{\pi_{\prior}(s_{t+1}\mid s_{t})}{\pi_{\phi}(s_{t+1}\mid s_{t})}.
\label{eq:trust-pcl-residual}
\end{equation}
This form suggests that Relative Trajectory Balance is exactly equivalent to Trust-PCL, in the sense that both losses are equal up to a constant factor that only depends on the temperature $\alpha$.

\begin{proposition}[Equivalence RTB -- Trust-PCL]
    \label{prop:equivalence-rtb-trust-pcl}
    The Relative Trajectory Balance loss \citep[RTB;][]{venkatraman2024rtb} defined with the residual in \cref{eq:relative-trajectory-balance-residual} is proportional to the Trust-PCL objective \citep{nachum2018trustpcl} defined with the residual in \cref{eq:trust-pcl-residual} on the MDP in \cref{sec:entropy-regularized-rl} (in particular, whose reward function satisfies \cref{eq:return-energy}): $\gL_{\trustpcl}(\phi, \psi) = \alpha^{2}\gL_{\mathrm{RTB}}(\phi, \psi)$, with
    \begin{align}
        V_{\soft}^{\psi}(s_{0}) &= \alpha \log Z_{\psi} &&& \pi_{\phi}(s'\mid s) &= P_{\phi}(s'\mid s)
        \label{eq:correspondance-rtb-trust-pcl}
    \end{align}
\end{proposition}

This result differs from similar ones relating GFlowNet objectives with MaxEnt-RL \citep{tiapkin2024gfnmaxentrl,deleu2024gfnmaxentrl} in that, unlike those existing connections, the equivalence in \cref{prop:equivalence-rtb-trust-pcl} does not require any correction of the reward function (with a backward transition probability). The proof of this proposition is immediate with the correspondence in \cref{eq:correspondance-rtb-trust-pcl}.

\subsection{Reinterpreting the empirical success of RTB}
\label{sec:reinterpreting-success-rtb}
Validating \cref{prop:equivalence-rtb-trust-pcl} empirically by comparing the performance of Trust-PCL against RTB would be of limited interest, since they both optimize the same loss (up to a constant factor). Therefore, we instead choose to revisit the results presented in the RTB paper in light of this equivalence. While \citet{venkatraman2024rtb} argued that ``\emph{RL methods with KL regularization yield inaccurate inference}'', our proposition suggests that this is not a fundamental issue of KL-regularized RL as a whole, and the success of RTB should be attributed to the choice of a superior RL algorithm, namely Trust-PCL.

\begin{figure}[t]
    \centering
    \begin{adjustbox}{center}
    \captionsetup{justification=centering}
    \begin{subfigure}[t]{0.25\textwidth}%
        \includegraphics[width=\linewidth]{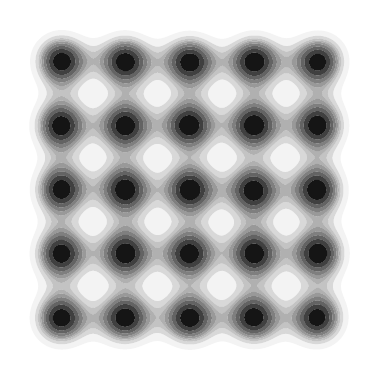}
        \caption{Prior $\pi_{\prior}^{\top}$}%
        \label{fig:25gmm-rtb-reinforce-comparison-0}%
    \end{subfigure}%
    \begin{subfigure}[t]{0.25\textwidth}%
        \includegraphics[width=\linewidth]{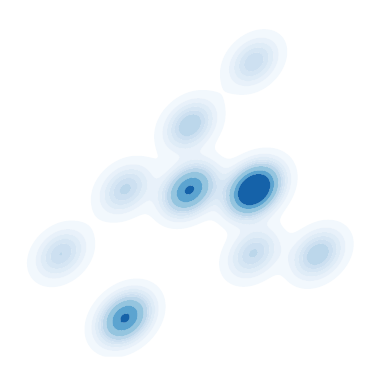}
        \caption{Target}%
        \label{fig:25gmm-rtb-reinforce-comparison-1}%
    \end{subfigure}%
    \begin{subfigure}[t]{0.25\textwidth}%
        \includegraphics[width=\linewidth]{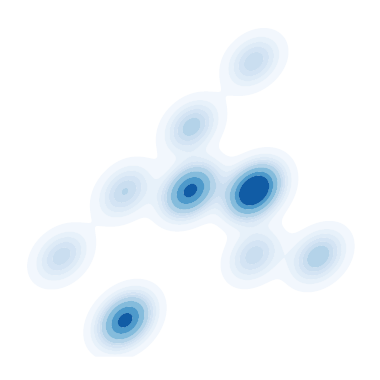}
        \caption{RTB}%
        \label{fig:25gmm-rtb-reinforce-comparison-2}%
    \end{subfigure}%
    \begin{subfigure}[t]{0.25\textwidth}%
        \includegraphics[width=\linewidth]{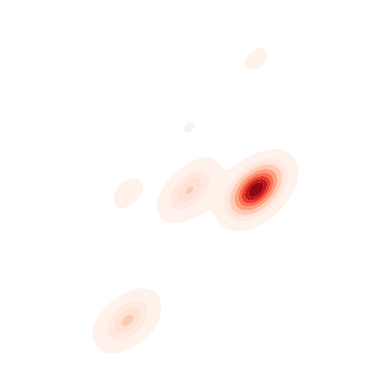}
        \caption{RL with KL reg.\newline (RTB paper)}%
        \label{fig:25gmm-rtb-reinforce-comparison-3}%
    \end{subfigure}%
    \begin{subfigure}[t]{0.25\textwidth}%
        \includegraphics[width=\linewidth]{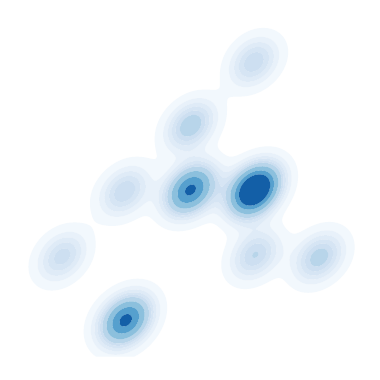}
        \caption{REINFORCE\newline with KL reg.}%
        \label{fig:25gmm-rtb-reinforce-comparison-4}%
    \end{subfigure}%
    \end{adjustbox}
    \caption{Comparison of RTB and RL with KL regularization on the illustrative example of \citep{venkatraman2024rtb}. (a) Prior distribution $\pi_{\prior}$. (b) Target tilted distribution $\propto \pi_{\prior}^{\top}(x)\exp(-\gE(x))$. (c) Model trained with RTB. (d) Model trained with REINFORCE with KL regularization, as reported by \citet{venkatraman2024rtb}. (e) Corrected model trained with off-policy REINFORCE with KL regularization.}
    \label{fig:25gmm-rtb-reinforce-comparison}
\end{figure}

To illustrate the apparent failure of RL with KL regularization for sequential generative modeling, \citet{venkatraman2024rtb} considered a 2D generation task where the prior is a uniform mixture of 25 Gaussians, and the target marginal distribution is a weighted mixture (\cref{fig:25gmm-rtb-reinforce-comparison-0,fig:25gmm-rtb-reinforce-comparison-1}). While RTB is capable of perfectly recovering the target distribution (\cref{fig:25gmm-rtb-reinforce-comparison-2}), they observed that RL with KL regularization seems to not be capturing all the modes of the target distribution (\cref{fig:25gmm-rtb-reinforce-comparison-3}).

At first glance, this could be explained by the difference in performance between their RL algorithm of choice, REINFORCE \citep{williams1992reinforce} and Trust-PCL/RTB. However, we show in \cref{app:off-policy-reinforce-kl} that this happens to be caused by their reward function not satisfying \cref{eq:return-energy}. This, in turn, changes the target distribution \cref{eq:tilted-optimal-kl-rl}, which does not match the one in \cref{fig:25gmm-rtb-reinforce-comparison-1} anymore. We show in \cref{fig:25gmm-rtb-reinforce-comparison-4} that if we apply the same method REINFORCE on the MDP of \cref{sec:entropy-regularized-rl} with minimal changes to ensure a fair comparison (\eg using off-policy data with self-normalized importance sampling correction \citep{precup2000eligibility}; see \cref{alg:off-policy-reinforce-kl} for details), then the expected target distribution can also be recovered as accurately as with RTB.

Going beyond this illustrative example, \citet{venkatraman2024rtb} also compared RTB \emph{on-policy} (\ie using $\pi_{b} \equiv P_{\phi}$ in \cref{sec:relative-trajectory-balance}) to existing methods based on RL for conditional image generation \citep{black2024ddpo,fan2023dpok}. Correcting for the discrepancy in the reward function mentioned above, we show in \cref{app:on-policy-rtb-reinforce} that in this setting, the gradient of RTB (on-policy) is equivalent to the one of REINFORCE with KL regularization, up to a different control variate. This mirrors existing results from the GFlowNet literature \citep{malkin2023gfnhvi}.

\section{Conclusion}
\label{sec:conclusion}
Reinforcement Learning with KL regularization has proven to be a versatile tool for fine-tuning sequential generative models. Its flexibility is such that even seemingly new methods for addressing the same problem can often be framed within this paradigm. In this work, we investigated the recently proposed Relative Trajectory Balance objective \citep{venkatraman2024rtb} through the lens of entropy-regularized Reinforcement Learning. We demonstrated that it is exactly equivalent to the Trust-PCL algorithm \citep{nachum2018trustpcl}, indicating that RTB effectively rephrases an existing KL-regularized RL approach within the GFlowNet framework. This is further evidence of the strong theoretical connections between these fields.

Leveraging this theoretical correspondence, we revisited the empirical claims by \citet{venkatraman2024rtb} regarding the shortcomings of KL-regularized RL. By carefully examining the illustrative experiment from the RTB paper, we found that simple methods such as REINFORCE with KL regularization can match RTB in modeling the target tilted distribution on that task. The reported failures stemmed from algorithmic and reward design choices rather than any fundamental limitation of Reinforcement Learning. However, on more challenging tasks, advanced methods like RTB/Trust-PCL are likely to offer clearer benefits, as suggested by recent work incorporating techniques such as update clipping \citep{fan2023dpok,black2024ddpo,shao2024deepseekmath} (inspired by PPO \citep{schulman2017ppo}). These observations emphasize the need for fair comparisons and well-posed problem formulations, while offering promising directions for capitalizing on the connections between GFlowNets and RL to explore new algorithms for generative model fine-tuning.

\newpage

\bibliography{references/references,references/gflownet}
\bibliographystyle{abbrvnat}


\newpage
\appendix
{\LARGE\textbf{Appendix}}

\section{Marginal distribution of the optimal KL-regularized policy}
\label{app:marginal-distribution-optimal-policy}
In this section, we prove that the marginal distribution associated with the optimal policy maximizing \cref{eq:kl-regularized-rl-objective} is the tilted distribution \cref{eq:tilted-optimal-kl-rl} for completeness, although this result can be found in various contexts in the literature \citep{nachum2018trustpcl, korbak2022rlwithklbayesian}. First, recall that the KL-regularized RL objective can be written as
\begin{equation}
    \pi^{\star}_{\mathrm{RelEnt}} = \argmax_{\pi} \E_{\tau \sim \pi}\left[\sum_{t=0}^{T}r(s_{t}, s_{t+1}) - \alpha \KL\big(\pi(\cdot\mid s_{t})\,\|\,\pi_{\prior}(\cdot\mid s_{t})\big)\right].
    \label{eq:kl-regularized-rl-objective-appendix}
\end{equation}
It can be shown that the optimal policy $\pi^{\star}_{\mathrm{RelEnt}}$ can be written in terms of a soft state-value function $V_{\soft}^{\star}(s)$ and a soft state-action value function $Q_{\soft}^{\star}(s, s')$ as
\begin{equation}
    \pi^{\star}_{\mathrm{RelEnt}}(s'\mid s) = \pi_{\prior}(s'\mid s)\exp\left(\frac{1}{\alpha}\big(Q_{\soft}^{\star}(s, s') - V_{\soft}^{\star}(s)\big)\right),
    \label{eq:optimal-policy-values}
\end{equation}
where the soft value functions, adapted to our setting where the MDP $\gM$ is finite horizon and deterministic, satisfy
\begin{align}
    Q_{\soft}^{\star}(s, s') &= r(s, s') + V_{\soft}^{\star}(s')\label{eq:soft-state-value}\\
    V_{\soft}^{\star}(s) &= \alpha \log \sum_{s'\in\mathrm{Ch}(s)}\pi_{\prior}(s'\mid s)\exp\left(\frac{1}{\alpha}Q_{\soft}^{\star}(s, s')\right),\label{eq:soft-state-action-value}
\end{align}
where $\mathrm{Ch}(s)$ are the children or the state $s$ in the MDP (\eg all the states of the form $(\vx_{t+1}, t+1)$ for a state $s_{t} = (\vx_{t}, t)$ in the example MDP given in \cref{sec:entropy-regularized-rl} \citep{fan2023dpok,black2024ddpo}). We defined the marginal distribution \cref{eq:terminating-state-distribution} in the main text for the particular example of diffusion samplers. However, this naturally generalizes in the context of KL-regularized RL, by viewing it as a marginal over trajectories in the MDP that terminate in $s_{T}$ (\ie the trajectory terminates with a transition $s_{T} \rightarrow \terminal$):
\begin{equation}
    \pi^{\top}(s_{T}) = \int\pi(\tau\mid s_{0})\mathbf{1}(s_{T}\rightarrow \terminal \in \tau)\,d\tau
    \label{eq:marginal-distribution-rl}
\end{equation}

For a trajectory $\tau = (s_{0}, s_{1}, \ldots, s_{T}, \terminal)$, we have
\begin{align}
    \pi^{\star}_{\mathrm{RelEnt}}(\tau\mid s_{0}) &= \prod_{t=0}^{T}\pi^{\star}_{\mathrm{RelEnt}}(s_{t+1}\mid s_{t})\label{eq:pi-trajectory-proof-1}\\
    &= \left[\prod_{t=0}^{T}\pi_{\prior}(s_{t+1}\mid s_{t})\right]\exp\left(\frac{1}{\alpha}\sum_{t=0}^{T}Q_{\soft}^{\star}(s_{t}, s_{t+1}) - V_{\soft}^{\star}(s_{t})\right)\label{eq:pi-trajectory-proof-2}\\
    &= \pi_{\prior}(\tau\mid s_{0})\exp\left(\frac{1}{\alpha}\sum_{t=0}^{T}r(s_{t}, s_{t+1}) + V_{\soft}^{\star}(s_{t+1}) - V_{\soft}^{\star}(s_{t})\right)\label{eq:pi-trajectory-proof-3}\\
    &= \pi_{\prior}(\tau\mid s_{0})\exp\left(\frac{1}{\alpha}(-\gE(s_{T}) + \cancel{V_{\soft}^{\star}(\terminal)} - V_{\soft}^{\star}(s_{0}))\right)\label{eq:pi-trajectory-proof-4}\\
    &\propto \pi_{\prior}(\tau\mid s_{0})\exp(-\gE(s_{T})/\alpha),\label{eq:pi-trajectory-proof-5}
\end{align}
where we used the definition of $\pi^{\star}_{\mathrm{RelEnt}}$ in \cref{eq:pi-trajectory-proof-2}, the definition of $Q_{\soft}^{\star}(s_{t}, s_{t+1})$ in \cref{eq:pi-trajectory-proof-3}, the property that the sum of reward functions only depends on the energy at the end of the trajectory \cref{eq:return-energy} and a telescoping sum in \cref{eq:pi-trajectory-proof-4}, and the fact that $V_{\soft}^{\star}(s_{0})$ is a constant independent of $\tau$ (and $V_{\soft}^{\star}(\terminal) = 0$) in \cref{eq:pi-trajectory-proof-5}. Plugging this in the definition of the marginal distribution \cref{eq:marginal-distribution-rl}, we get
\begin{align}
    \pi_{\mathrm{RelEnt}}^{\star\top}(s_{T}) &= \int \pi_{\mathrm{RelEnt}}^{\star}(\tau\mid s_{0})\mathbf{1}(s_{T}\rightarrow \terminal \in \tau)\,d\tau\\
    &\propto \exp(-\gE(s_{T})/\alpha)\int \pi_{\prior}(\tau\mid s_{0})\mathbf{1}(s_{T}\rightarrow \terminal \in \tau)\,d\tau\\
    &\propto \pi_{\prior}^{\top}(s_{T})\exp(-\gE(s_{T})/\alpha).
\end{align}

\section{Off-policy REINFORCE with KL regularization}
\label{app:off-policy-reinforce-kl}
We saw in \cref{sec:entropy-regularized-rl} that the marginal distribution associated with the optimal policy $\pi_{\mathrm{RelEnt}}^{\star}$ matches the tilted distribution \cref{eq:tilted-optimal-kl-rl} in the case where the reward function of the MDP satisfies
\begin{equation}
    \sum_{t=0}^{T}r(s_{t}, s_{t+1}) = -\gE(s_{T}).
\end{equation}
In their comparison with KL-regularized RL, \citet{venkatraman2024rtb} instead used a sparse reward function that equals $\tilde{r}(s_{T}, \terminal) = \exp(-\gE(s_{T}))$ at the terminating transition, and zero everywhere else, meaning that their optimum would be
\begin{equation}
    \pi^{\star\top}(s_{T}) \propto \pi_{\prior}^{\top}(s_{T})\exp(\exp(-\gE(s_{T}))/\alpha),
\end{equation}
where we emphasize the extra inner ``$\exp$''. This error likely came from the naming conflict between the ``reward function'' in RL and the ``reward'' in the GFlowNet literature, where the latter should indeed equal the exponential of the corresponding reward in RL \citep{deleu2024gfnmaxentrl}. This can be seen in the residual for RTB \cref{eq:relative-trajectory-balance-residual}, which can be rewritten as
\begin{equation}
    \Delta_{\mathrm{RTB}}(\tau; \phi, \psi) = \log \frac{\overbrace{\exp(-\gE(s_{T})/\alpha)}^{\mathclap{\textrm{GFlowNet reward ``$R(s_{T})$''}}}\prod_{t=0}^{T}\pi_{\prior}(s_{t+1}\mid s_{t})}{Z_{\psi}\prod_{t=0}^{T}P_{\phi}(s_{t+1}\mid s_{t})}.
\end{equation}

\begin{figure}[t]
\begin{adjustbox}{center}
\begin{minipage}{0.6\linewidth}
\begin{algorithm}[H]
    \caption{RL with KL reg. \citep{venkatraman2024rtb}}
    \label{alg:on-policy-reinforce}
    \begin{algorithmic}[1]
        \State Sample a batch of trajectories $\{\tau_{1}, \ldots, \tau_{N}\}$ using $\pi_{\phi}$
        \State Compute the cumulative reward for $\tau_{n}$\vphantom{$\frac{1}{N}\sum_{n=1}^{N}$}:
        \vspace{-0.5em}
        \begin{align*}r_{n} &\leftarrow \exp(-\gE(s_{T}^{(n)}))\vphantom{-\alpha\log \frac{\pi_{\phi}(\tau_{n})}{\pi_{\prior}(\tau_{n})}}\end{align*}
        \vspace{-0.5em}
        \vspace{0.5em}
        \State Compute the advantage: $\bar{r}_{n} = r_{n} - \frac{1}{N}\sum_{n=1}^{N}r_{n}$
        \addtocounter{ALG@line}{1}
        \Statex \vphantom{Get the SNIS weights: $w_{n} \propto \big(\pi_{\phi}(\tau_{n}) / \pi_{b}(\tau_{n})\big)^{1/T}$}
        \State Compute the REINFORCE loss with KL reg.\vphantom{$\frac{1}{N}\sum_{n=1}^{N}$}
        \vspace{-0.5em}
        \begin{align*}
        \hspace*{-1.7em}
            \gL(\phi) &= \frac{1}{N}\sum_{n=1}^{N} -\mathrm{sg}(\bar{r}_{n})\log \pi_{\phi}(\tau_{n}) + \frac{\lambda}{2} \left(\log \frac{\pi_{\phi}(\tau_{n})}{\pi_{\prior}(\tau_{n})}\right)^{2}
        \end{align*}
    \end{algorithmic}
\end{algorithm}
\end{minipage}
\hspace{0.5em}%
\begin{minipage}{0.6\linewidth}
\begin{algorithm}[H]
    \caption{Off-policy REINFORCE with KL reg.}
    \label{alg:off-policy-reinforce-kl}
    \begin{algorithmic}[1]
        \State Sample a batch of trajectories $\{\tau_{1}, \ldots, \tau_{N}\}$ using $\pi_{b}$
        \State Compute the cumulative reward for $\tau_{n}$\vphantom{$\frac{1}{N}\sum_{n=1}^{N}$}:
        \vspace{-0.5em}
        \begin{align*}r_{n} &\leftarrow -\gE(s_{T}^{(n)})-\alpha\log \frac{\pi_{\phi}(\tau_{n})}{\pi_{\prior}(\tau_{n})}\end{align*}
        \vspace{-0.5em}
        \vspace{0.5em}
        \State Compute the advantage: $\bar{r}_{n} = r_{n} - \frac{1}{N}\sum_{n=1}^{N}r_{n}$
        \State Get the SNIS weights: $w_{n} \propto \big(\pi_{\phi}(\tau_{n}) / \pi_{b}(\tau_{n})\big)^{1/T}$
        \State Compute the REINFORCE loss\vphantom{$\frac{1}{N}\sum_{n=1}^{N}$}
        \vspace{-0.5em}
        \begin{align*}
            \gL(\phi) &= -\frac{1}{N}\sum_{n=1}^{N} \mathrm{sg}(w_{n}\bar{r}_{n})\log \pi_{\phi}(\tau_{n})
        \end{align*}
    \end{algorithmic}
\end{algorithm}
\end{minipage}
\end{adjustbox}
\end{figure}

Besides this mismatch in reward functions, we made some minor updates to the REINFORCE \citep{williams1992reinforce} algorithm from \cref{alg:on-policy-reinforce} considered in the RTB paper to \cref{alg:off-policy-reinforce-kl} (``$\mathrm{sg}$'' is the stop-gradient operator), to ensure a fair comparison with RTB:
\begin{itemize}[noitemsep, topsep=1ex, itemsep=1ex, leftmargin=2em]
    \item The trajectories are collected using the same exploratory behavior policy $\pi_{b}$ as RTB, to address the well-known limitations of on-policy methods in terms of exploration (this was also noted by \citet{venkatraman2024rtb}). To that end, we use importance sampling to address this shift in distributions (between $\pi_{b}$ and the policy $\pi_{\phi}$ being learned), and more particularly self-normalized importance sampling \citep[SNIS;][]{precup2000eligibility} to control the variance of the gradient estimate.
    \item The log-ratio $\log \pi_{\phi}/\pi_{\prior}$ is added to the reward itself, as opposed to being a squared regularization, so that it can be incorporated into the average control variate. Note that while \citet{venkatraman2024rtb} treated the regularization constant $\lambda$ as a separate hyperparameter, they had to tune carefully (with $\lambda = \alpha = 1$ being the best value they found), we saw in \cref{sec:entropy-regularized-rl} that this exactly corresponds to the temperature parameter $\alpha$ of the tilted distribution.
\end{itemize}
Going beyond the illustrative example of \cref{fig:25gmm-rtb-reinforce-comparison}, we note that in the code release accompanying the paper, \citet{venkatraman2024rtb} only considered baselines based on vanilla REINFORCE in their comparisons with KL-regularized RL methods. However, in practice, methods such as DDPO \citep{black2024ddpo} \& DPOK \citep{fan2023dpok} (referenced as RL baselines) typically incorporate additional techniques, including update clipping.

\section{On-policy RTB is equivalent to REINFORCE with KL regularization}
\label{app:on-policy-rtb-reinforce}
In this section, we will show that the gradient of on-policy RTB is equal to the gradient of REINFORCE with KL regularization, up to a constant $\alpha$. First, recall that the gradient of the REINFORCE loss with KL regularization can be written as
\begin{equation}
    \nabla_{\phi}\gL_{\mathrm{RL}}(\phi) = -\E_{\pi_{\phi}}\left[\left(-\gE(s_{T}) - \alpha \log \frac{\pi_{\phi}(\tau)}{\pi_{\prior}(\tau)} - b\right)\nabla_{\phi}\log \pi_{\phi}(\tau)\right],
    \label{eq:gradient-reinforce}
\end{equation}
where $b$ is a control variate (baseline), which is often estimated using an average over the batch of trajectories (see \cref{alg:off-policy-reinforce-kl}). Similarly, recall that the on-policy (\ie where $\pi_{b} \equiv \pi_{\phi}$) RTB loss in \cref{eq:relative-trajectory-balance-residual} can be written as
\begin{align}
    \gL_{\mathrm{RTB}}(\phi, \psi) &= \frac{1}{2}\E_{\pi_{\phi}}\left[\left(\log \frac{\pi_{\prior}(\tau)}{Z_{\psi}\pi_{\phi}(\tau)} - \frac{\gE(s_{T})}{\alpha}\right)^{2}\right],
\end{align}
where we use the notation $\pi_{\phi}$ instead of $P_{\phi}$ to match the policy in KL-regularized RL. Taking the gradient of $\gL_{\mathrm{RTB}}$ wrt. $\phi$, and ignoring differentiation through the policy over which we take the expectation (which is standard in the GFlowNet literature \citep{malkin2023gfnhvi}), we get
\begin{align}
    \nabla_{\phi}\gL_{\mathrm{RTB}}(\phi, \psi) &= \frac{1}{2}\E_{\pi_{\phi}}[\nabla_{\phi}\Delta_{\mathrm{RTB}}^{2}(\tau; \phi, \psi)]\\
    &= \E_{\pi_{\phi}}\left[\left(\frac{\gE(s_{T})}{\alpha} + \log \frac{\pi_{\phi}(\tau)}{\pi_{\prior}(\tau)} + \log Z_{\psi}\right)\nabla_{\phi}\log \pi_{\phi}(\tau)\right]\\
    &= \frac{1}{\alpha}\nabla_{\phi}\gL_{\mathrm{RL}}(\phi),
\end{align}
if the baseline $b = \alpha \log Z_{\psi}$. Therefore, both methods are equivalent, only differing in their control variate: REINFORCE with KL regularization often use a local baseline estimated using the current batch of trajectories, whereas $Z_{\psi}$ in on-policy RTB acts as a \emph{global} baseline. This result is similar to the one derived by \citet{malkin2023gfnhvi}.

\end{document}